\documentclass[]{article}
\usepackage{proceed2e}
\usepackage{natbib}
\usepackage{graphicx}
\usepackage{verbatim}
\usepackage{url}
\usepackage{subfigure}
\usepackage{algorithm, algorithmic}
\usepackage{amsmath, amsthm, amsfonts, amssymb}
\usepackage[usenames]{color}
\usepackage{tikz}
\usepackage{booktabs}
\usepackage{multirow}
\usepackage{rotating}
\usepackage{bm}
\usepackage{wrapfig}
\usepackage{multicol}
\usepackage{epsfig}

\title{Communication: Words and Conceptual Systems}

\author{ {\bf Jian YU } \\
Beijing Key Lab
of Traffic Data Analysis and Mining\\
 Beijing Jiaotong University,Beijing, China\\
Email: jianyu@bjtu.edu.cn
}

\begin{document}
%\begin{CJK*}{GBK}{song}

\maketitle

\begin{abstract}%   <- trailing '%' for backward compatibility of .sty file
 Words (phrases or symbols) play a key role in human life. Word (phrase or symbol) representation is the fundamental problem for knowledge representation and understanding. A word (phrase or symbol) usually represents a name of a category. However, it is always a challenge that how to represent a category can make it easily understood.  In this paper, a new representation for a category is discussed, which can be considered a generalization of classic set.  In order to reduce representation complexity, the economy principle of category representation is proposed. The proposed category representation provides a powerful tool for analyzing conceptual systems, relations between words, communication, knowledge, situations. More specifically,  the conceptual system, word relations and communication are mathematically defined and classified such as ideal conceptual system, perfect communication and so on; relation between words and sentences is also studied, which shows that knowledge are words. Furthermore, how conceptual systems and words depend on situations is presented, and how truth is defined is also discussed.
\end{abstract}

{\bf Keywords:} Word, Category, Conceptual System, Relations between Words, Sentence, Communication, Knowledge, Situation, Truth

\section{Introduction}

When studying objects in the real or virtual world, categorization is a common approach in order to communicate with each other. A word or phrase (sometimes, symbol) represents the corresponding category name. Every minute in daily life, we rely on words (phrases or symbols) to help us to deal with everything, such as talking, listening, watching, thinking, etc. Transparently, a word (phrase or symbol) usually represents a name of a category. How to represent a category becomes a pivotal question in word understanding. In the literature, a classical category has three representation: the corresponding name representation, the corresponding set representation, the corresponding proposition representation. In common sense, a word or phrase represent the name of a category, which is also called the concept name or the category name. Set representation represents a class of objects in the world belonging to the corresponding category, proposition representation represents the mental representation of the corresponding category.

Up to now, all the above three representations seem in question. Firstly, a category usually has different names in different languages. What's more, a category  has different names in the same language. For example, when seeing a specific dog, we can say it ``dog'' in English, ``¹·'' in Chinese. Clearly, the word ``dog''  is different from that specific dog in the world, which only is the name of the category ``dog''. Hence, the name of a category cannot fully reflect the meaning of this category. In extremity, the name of a category is only considered as a symbol and has no meaning.  Furthermore,  sometimes man cannot find the right name of a category for what he wants to say, and sometimes man does not like to say the right name of a category for what he should say. In other words, a category may have different name in the mind from  that in the utterance.   \cite{DePaulo1996} have stated that lying was an everyday event, and \cite{Heyman2009} have discovered that parents lie to their children even through they maintain that lying is unacceptable.   Secondly, a category may be very fuzzy or vague such as beauty, badness and so on. Clearly, a classical set can not express this case.  More surprised,  \cite{watanabe1969knowing} has stated that it is impossible to distinguish a category from others without feature selection. Thirdly, \cite{wittgenstein1953philosophical} has claimed that important categories such as games may not be defined (represented) by a proposition.  Such view has been widely accepted in cognitive science fields \citep{lako1987women}. Fourthly, it is taken for granted that all the men should have the same representation for a classical category. What a pity, such an assumption is also clearly not correct, otherwise, misunderstanding about the same category cannot occur among men in daily life.  Cognition research results have proved that most concepts are not disembodied but  embodied \citep{lako1987women}.  In the final,  it is usually supposed that set representation is equivalent to proposition representation with respect to categorization.  However, such an assumption may be false in daily life. %In China, it is well known that `` �Բ�����'' as early as the Warring States Period~\citep{zhouyi}.

  %How to distinguish concepts from categories? How to represent them?

Although there exist the above drawbacks, man still can use words (phrases or symbols) to express their thoughts, feelings and observations. Why?  In this paper, we devote to  answering this question and make major contributions as follows. %an object $x$ in the discussed domain is absolutely in or not in the corresponding category according to the similarity between the object $x$ and the discussed concept representation.  In other words, the  membership function is 0 or 1 function, but the corresponding similarity  between the object $x$ and the category is continuous in $[0,1]$. Clearly, similarity is not equal to membership.

1) Semantic set is proposed to represent a category in Section 2.

2) Conceptual system is discussed in Section 3.

3) Relations between words  are studied in Section 4.

4) How to communicate concepts between people is discussed in Section 5.

5) Relation between words and sentences is discussed in Section 6.

6) How conceptual systems and words depend on situations is presented in Section 7.

7) The relation between semantic set and its referent is discussed in Section 8.

8) The definitions of truth and falsity are investigated in Section 9.

\section{Category and Semantic Set}

It is well known in China that The Named is the mother of all things\citep{Laotzu1979} and the word is used to refer to its corresponding objects in the world \citep{Gongsunlongzi}, which usually is limited in a specific domain.  Therefore, let the discussed object domain be a set $O$ and $o$ represent an object in $O$, the discussed category domain be a conceptual system $L$ and $\mathfrak{A}$ denotes a category  in $L$.  For different categories in $L$,  $o$ may have different representations. For example, when categorization by obesity,  a man can be represented by his age,  his weight, his height and so on; when categorization of different types of individuals, a man can be represented by his sensation, his intuition, his feeling,his thinking,his attitudes and so on. Transparently,  the object representation depends on categorization. Ugly Duckling Theorem \citep{watanabe1969knowing} has shown that a category specializes its object representation, otherwise, objects cannot be distinguished by different categories.
In this paper, when specific categorization is needed, the corresponding object representation is supposed to have been obtained. Therefore, let $\mathfrak{A_O}$ be the category name we will discuss, $O_A=\{o_A| o\in O\}$, where $o_A$ is the object representation of the object $o$ in the set $O$ with respect to the category $\mathfrak{A}$,   $I_A$ is the corresponding membership function defined as follows. %As \cite{Betrandrusell1950},

{\bf Membership function:} \\
  $I_A$: $O_A\mapsto R_{+}$ is called the membership function of $\mathfrak{A}$ if $\forall o_A\in O_A$, the greater $I_A(o_A)$ means the more probability $o$ belongs to the category  $\mathfrak{A}$.

As $O_A$ and $I_A$ are observable,  $\{O_A,I_A\}$ is called the outer set of $\mathfrak{A}$. If $\forall o\in O, I_A(o_A)\in \{0,1\}$, then it is easy to know that $\forall o$,  $I_A(o_A)=1$  if and only if $o$ belongs to the category $\mathfrak{A}$,  and $I_A(o_A)=0$ if and only if $o$ does not belong to the category $\mathfrak{A}$.  In this case,  $\{O_A,I_A\}$ represents a classic set. If $\forall o\in O, I_A(o_A)\in [0,1]$, then $\{O_A,I_A\}$ represents a fuzzy set\citep{Zadeh1965fuzzyset}.

In order to understand a category, it is reasonable to suppose that any category should have its corresponding concept in mental space \citep{JianYu2015Categorization}. Therefore, let $\underline{A}$ represent the concept corresponding to the category $\mathfrak{A}$ and the name of $\underline{A}$ be $\mathfrak{A_I}$. When the concept representation for any category is defined, objects can be categorized based on the similarity between objects and the concept representation. The  category similarity mapping  can be defined by computing the similarity between objects and  the concept representation as follows.

 {\bf Category Similarity Mapping:} \\
  $Sim_A$: $O_A\times \underline{A} \mapsto R_{+}$ is called the category similarity mapping of $\mathfrak{A}$ if $\forall o\in O$, the increase of $Sim_A(o_A,\underline{A})$ means the greater similarity between the object $o$ and the category $\mathfrak{A}$,   where $o_A$ is the object representation with respect to $\mathfrak{A}$.

 Similarly, the category dissimilarity mapping  can be defined as follows:

  {\bf Category Dissimilarity Mapping:} \\
  $Ds_A$: $O_A\times \underline{A} \mapsto R_{+}$ is called  the category dissimilarity mapping  of $\mathfrak{A}$ if  $\forall o\in O$, the decrease of $Ds_A(o_A,\underline{A})$ means the greater similarity between the object $o$ and the category $\mathfrak{A}$,   where $o_A$ is the object representation with respect to $\mathfrak{A}$.

Generally speaking, $(\underline{A},Sim_A)$ or $(\underline{A},Ds_A)$ can be unobservable, therefore, $(\underline{A},Sim_A)$ or $(\underline{A},Ds_A)$ is called the inner set of the category $\mathfrak{A}$, its name is called $\mathfrak{A_I}$.

For brevity, ($\mathfrak{A_O},O_A,I_A$) is called the outer representation of the category $\mathfrak{A}$,($\mathfrak{A_I},\underline{A},Sim_A$) is called the inner representation of the category $\mathfrak{A}$.  In language, $\mathfrak{A_O}$ is the outer name of the category, $\underline{A}$ is the mental representation of $\mathfrak{A}$, $\mathfrak{A_I}$ is the inner name of of the category. Usually, $\mathfrak{A_O}$ corresponds to a word  (phrase or symbol) in a specific language, $\mathfrak{A_I}$ also corresponds to a word (phrase or symbol) in a specific language.  In summary, a category $\mathfrak{A}$ can be represented by a six tuple: $(\mathfrak{A_O},O_A,I_A,\mathfrak{A_I},\underline{A},Sim_A)$ or $(\mathfrak{A_O},O_A,I_A,\mathfrak{A_I},\underline{A},Ds_A)$.  $(\mathfrak{A_O},O_A,I_A,\mathfrak{A_I},\underline{A},Sim_A)$ or $(\mathfrak{A_O},O_A,I_A,\mathfrak{A_I},\underline{A},Ds_A)$ is called a semantic set.  Usually, a semantic set depends on its user.  Hence, a semantic set is a generalization of a classic set, which overcomes the drawbacks of the classical category representation and accords with the view of point in \citep{LakofeJohnson1980}. %If $\underline{A}=P_A()$ and $\forall o_A, I_A(o_A)=Sim_A(o_A,\underline{A})$ and $\mathfrak{A_O}=\mathfrak{A_I}$, then the semantic set $(\mathfrak{A_O},O_A,I_A,\mathfrak{A_I},\underline{A},Sim_A)$ can be reduced to $(\mathfrak{A_O},O_A,I_A)$.

%In daily life, a category is often represented by its category name. When a man speaks some category, he has its outer representation and its inner representation. Therefore, a category  %Transparently, when $A\neq A_I$, the speaker does not speak the truth and violates the cooperative principle in conversation, especially the maxim of quality, which states that do not say what you believe to be false.

In daily life, a category $\mathfrak{A}$ is often represented by its outer name $\mathfrak{A_O}$. However, only the outer name of the category is known, it will be a puzzle for man to understand this category as the same name sometimes refers to many categories.  Many words (phrases or symbols) in ancient classics cannot be understood by modern man just because their corresponding outer sets and inner sets have been lost in the long river of history.

\section{Conceptual System}

%For a category $\mathfrak{A}$, its semantic set is ($A,I_A,\underline{A},Sim_A$).

As \cite{lako1987women} has pointed out, our conceptual systems grow out of bodily experience; moreover, the core of our conceptual systems is directly grounded in perception, body movement, and experience of a physical and social character.  Therefore, the meaning of any word depends on a man's physical and social environments. Transparently, different men may have different category representations with respect to the same category $\mathfrak{A}$.

\par Every man has at least one conceptual system, which includes many categories.  According the above analysis,  the man $\alpha$'s conceptual system can be defined as $L^{\alpha}=\{\mathfrak{A}^{\alpha}\}$, where the category $\mathfrak{A}^{\alpha}$ can be represented by $(\mathfrak{A_O^{\alpha}},O_A^{\alpha},I_A^{\alpha},\mathfrak{A_I^{\alpha}},\underline{A}^{\alpha},Sim_A^{\alpha})$. 
 $\mathfrak{A_O^{\alpha}}$ is a word  (phrase or symbol),  represents the referring action with respect to $\mathfrak{A}^{\alpha}$, $(\mathfrak{A_I^{\alpha}},\underline{A}^{\alpha},Sim_A^{\alpha})$ represents the mental representation concerned on the category $\mathfrak{A}^{\alpha}$. If the above three parts of the category $\mathfrak{A}^{\alpha}$ are not contradictory,  then $\mathfrak{A^{\alpha}}$ is self-consistent.
How to say a category $\mathfrak{A^{\alpha}}$ is self-consistent?

\par In order to answer this question,  the inner referring operator $\sim$ can be defined as follows:
\begin{equation}\label{innerreferring}
 \widetilde{o_A^{\alpha}}=  \arg\max_{\mathfrak{B_I^{\alpha}}}Sim_{B}^{\alpha}(o_B^{\alpha},\underline{B}^{\alpha})
\end{equation}
where $o_A^{\alpha}=o_B^{\alpha}, \mathfrak{B^{\alpha}}\in L^{\alpha}$;

The outer referring operator $\rightarrow$ can be defined as follows:
\begin{equation}\label{outerreferring}
\overrightarrow{o_A^{\alpha}}=\arg\max_{\mathfrak{B_O^{\alpha}}}I_{B}^{\alpha}(o_B^{\alpha})
\end{equation}
 where $o_A^{\alpha}=o_B^{\alpha}$, $\mathfrak{B^{\alpha}}\in L^{\alpha}$.

Sometimes, $\overrightarrow{o_A^{\alpha}}$ or $\widetilde{o_A^{\alpha}}$ is multiple value.  When $\overrightarrow{o_A^{\alpha}}$ or $\widetilde{o_A^{\alpha}}$ is multiple value, it is very difficult for man $\alpha$ to recognize the object $o$.

If $\forall o\in O (\overrightarrow{o_A^{\alpha}} $ and $\widetilde{o_A^{\alpha}}$ are single value), it is natural to define that
$A_O^{\alpha}=\{o\in O|\overrightarrow{o_A^{\alpha}}=\mathfrak{A_O^{\alpha}}\}$ and $A_I^{\alpha}=\{o\in O|\widetilde{o_A^{\alpha}}=\mathfrak{A_I^{\alpha}}\}$, where $A_O^{\alpha}$ is called the outer referring set and $A_I^{\alpha}$  the inner referring set with respect to the category $\mathfrak{A^{\alpha}}$. If $o\in A_O^{\alpha}$, then $o$ can be externally called $\mathfrak{A_O^{\alpha}}$, in other words, $o$ has an outer name $\mathfrak{A_O^{\alpha}}$. If $o\in A_I^{\alpha}$, then $o$ can be internally called $\mathfrak{A_I^{\alpha}}$, in other words, $o$ has an inner name $\mathfrak{A_I^{\alpha}}$. In daily life, sometimes man can call black into white. Hence it is not always true that $\mathfrak{A_I^{\alpha}}$=$\mathfrak{A_O^{\alpha}}$ for the man $\alpha$ with respect to one object $o\in O$.  If $A_O^{\alpha}=A_I^{\alpha}$ and $\mathfrak{A_O^{\alpha}}=\mathfrak{A_I^{\alpha}}$, then $\mathfrak{A}^{\alpha}$ is self-consistent.

 \par Self-consistent is very helpful to reduce the complexity of a semantic set. If a semantic set is self-consistent, it is the simplest because its inner representation can be ignored in some sense. Therefore, self-consistent is the economy assumption for category representation.  However, self-consistent is not always true for a conceptual system.  If  $\forall \mathfrak{A^{\alpha}}\in L^{\alpha}$ such that $A_O^{\alpha}=A_I^{\alpha}$ ,
 there exist the mappings between the inner name and the outer name as follows.

 {\bf Name Encoding Mapping:} \\
  $N_e^{\alpha}$: $ \mathfrak{A_O^{\alpha}}\mapsto \mathfrak{A_I^{\alpha}}$ is called the name encoding mapping of $L^{\alpha}$, where  $\forall \mathfrak{A^{\alpha}} \in L^{\alpha}$ such that $A_O^{\alpha}=A_I^{\alpha}$ .

  {\bf Name Decoding Mapping:} \\
  $N_d^{\alpha}$: $ \mathfrak{A_I^{\alpha}}\mapsto \mathfrak{A_O^{\alpha}}$ is called the name decoding mapping of $L^{\alpha}$, where  $\forall \mathfrak{A^{\alpha}}\in L^{\alpha}$ such that $A_O^{\alpha}=A_I^{\alpha}$.

%{\bf Referring Encoding Mapping:} \\
 % $R_e^{\alpha}$: $ \widetilde{o_A^{\alpha}} \mapsto \overrightarrow{o_A^{\alpha}}$ is called the name encoding mapping of $L^{\alpha}$, where $\forall o\in O$ and $\forall \mathfrak{A^{\alpha}}\in L^{\alpha}$.

%{\bf Referring decoding Mapping:} \\
%  $R_d^{\alpha}$: $ \overrightarrow{o_A^{\alpha}}\mapsto \widetilde{o_A^{\alpha}}$ is called the name decoding mapping of $L^{\alpha}$, where $\forall o\in O$ and $\forall \mathfrak{A^{\alpha}}\in L^{\alpha}$.

According to the above analysis,  a conceptual system can be accurately expressed as $L^{\alpha}_{\{N_e^{\alpha},N_d^{\alpha}\}}=\{\mathfrak{A}^{\alpha}\}$ if $\forall \mathfrak{A^{\alpha}}\in L^{\alpha}$ such that $A_O^{\alpha}=A_I^{\alpha}$ .  A conceptual system $L^{\alpha}_{\{N_e^{\alpha},N_d^{\alpha}\}}$ is called a plain conceptual system. When its name encoding mapping and its name decoding mapping are an identity function, it can be simply denoted by $L^{\alpha}_I$, such a conceptual system is called a self-consistent conceptual system.

For a category in the conceptual system $L^{\alpha}_{\{N_e^{\alpha},N_d^{\alpha}\}}\neq L^{\alpha}_I$, it can not be guaranteed to be self-consistent.  In this case, it is more difficult to understand the conceptual system $L^{\alpha}_{\{N_e^{\alpha},N_d^{\alpha}\}}$ than to understand the conceptual system $L^{\alpha}_I$.  In daily life, translation among different languages can be considered to seek the optimal name encoding mapping and name decoding mapping.  Certainly, when the mappings between the inner name and the outer name do not exist, the inner referring set and the outer referring set need to be calculated independently, hence, the corresponding conceptual system is the most difficult to be understood.

When the name encoding mapping and the name decoding mapping do exist for a conceptual system, it is enough to compute either the inner referring set or the outer referring set as the inner referring set is equal to the outer referring set.  In order to further reduce the difficulty of understanding, it is natural to require that the name encoding mapping and the name decoding mapping are an identity function, which means such a conceptual system is self-consistent.  Moreover, if any concept can be defined by a proposition in a self-consistent conceptual system, such a conceptual system is called an ideal conceptual system.  Certainly, an ideal conceptual system can be accurately expressed by natural language. If a conceptual system is not ideal, it may not be accurately expressed by a natural language. Therefore, an ideal conceptual system is the most easily understood and inherited. However, it is the most difficult for man to obtain an ideal conceptual system.  In general,  man only can continuously approximate an ideal conceptual system through making $\underline{A}$ be expressed by some proposition as accurately and precisely as possible for any category $\mathfrak{A}$ in his conceptual system.

\section{Relations Between Words}

 A conceptual system $L$ has many words, its two words usually have complex relations.  In order to simplify to study relations among words, two assumptions are made as follows:
   \par  \textbf{1): $L$ is self-consistent.}
   \par \textbf{2): $\forall o\in O$, $\forall \mathfrak{A}\in L$, $\forall \mathfrak{B}\in L$, $o_A$ can be expressed by a vector with fixed length $l$ such as $[o_A^{1},o_A^{2},\cdots,o_A^{l}]$, $o_B$ can be expressed by a vector with fixed length $m$ such as $[o_B^{1},o_B^{2},\cdots,o_B^{m}]$,where $o_A^{i}$ has a feature name $A^f_i$ in $L$, and $A^f=\{A^f_1,A^f_2,\cdots,A^f_l\}$ is the feature set of the category $\mathfrak{A}$, $o_B^{i}$ has a feature name $B^f_i$ in $L$, and $B^f=\{B^f_1,B^f_2,\cdots,B^f_m\}$ is the feature set of the category $\mathfrak{B}$.
 }

 According to the above two assumptions,  $\forall o\in O$,  $o_A$ can be defined as $[A^f_1(o),A^f_2(o),\cdots,A^f_l(o)]$,   $o_B$ can be defined as $[B^f_1(o),B^f_2(o),\cdots,B^f_m(o)]$,  where $A^f_1(o),A^f_2(o),\cdots, A^f_l(o),B^f_1(o),B^f_2(o),\cdots,B^f_m(o)$ can be considered to be functions.

 If $\mathfrak{A_O}$=$\mathfrak{B_O}$ and $A_O\bigcap B_O=\emptyset$ and $A^f\bigcap B^f=\emptyset$, then the word  $\mathfrak{A_O}$ and the word $\mathfrak{B_O}$ are called homonymy, which have the same category name but with different meanings. Theoretically, every word is  homonymy as every word can be  self-referring or non self-referring.  For example, the word ``dog'' can refer to not only the word ``dog'' itself  but also  a real dog in the the world.

 If $\mathfrak{A_O}$=$\mathfrak{B_O}$ and $(A_O \bigcap B_O\neq \emptyset)\bigvee (\exists i \exists j (A^f_i=B^f_j))$ and $A_O \neq B_O$,  then the word  $\mathfrak{A_O}$ and the word $\mathfrak{B_O}$ are called  polysemy.

 If $\mathfrak{A_O}\neq \mathfrak{B_O}$ and $A_O=B_O$, then the word  $\mathfrak{A_O}$ and the word $\mathfrak{B_O}$ are called synonymy, which have the different category names but with the same meaning.

 If $\mathfrak{A_O}\neq \mathfrak{B_O}$ and $A_O \subset B_O$, then the word  $\mathfrak{A_O}$ and the word $\mathfrak{B_O}$ are called hyponymy. More detailed,  the word  $\mathfrak{A_O}$ is called the hyponym of  the word  $\mathfrak{B_O}$,  the word  $\mathfrak{B_O}$ is called the hypernym of  the word  $\mathfrak{A_O}$.

 If $\mathfrak{A_O}\neq \mathfrak{B_O}$ and $A_f = B_f$ and $A_O \cap B_O=\emptyset$, then the word  $\mathfrak{A_O}$ and the word $\mathfrak{B_O}$ are called antonymy.

  If $\mathfrak{A_O}\neq \mathfrak{B_O}$ and $\forall o\in A_O \exists \dot{o}\in B_O (o\in \dot{o})$, then the word  $\mathfrak{A_O}$ and the word $\mathfrak{B_O}$ are called meronymy,  the word  $\mathfrak{A_O}$ is called the meronym of  the word  $\mathfrak{B_O}$,  the word  $\mathfrak{B_O}$ is called the holonym of  the word  $\mathfrak{A_O}$.

  If $\mathfrak{A_O}\neq \mathfrak{B_O}$ and $\mathfrak{A_O}\in B_O $, then the word $\mathfrak{B_O}$ is said to modify the word $\mathfrak{A_O}$. Usually, the word $\mathfrak{A_O}$ is called the modified word, the word $\mathfrak{B_O}$ is called the modifier.

  Sometimes, more complex relations of two words $\mathfrak{A_O}$ and $\mathfrak{B_O}$ are needed to be considered.  For example, metaphor and metonymy also illustrate relation between two words that seem irrelevant.   In theory, if $\mathfrak{A_O}\neq \mathfrak{B_O}$ and $ \exists i \exists j (A^f_i=B^f_j)$, then one can directly use the word $\mathfrak{A_O}$  to refer to the category $\mathfrak{B}$, or one can say that $\mathfrak{A_O}$ is $\mathfrak{B_O}$ in order to make description simple, vivid and accurate. Here, it should be pointed out that category feature mapping usually depends on the situation, which includes relevant people, places, times, objects and environments.  %Generally speaking, category feature mapping depends on situations in reality although it is expected to be independent of situations in ideal circumstances.

  In many applications, it is very important to compute the semantic similarity between two words.  According to the proposed category representation, categories $\mathfrak{A}$ and $\mathfrak{B}$ are said to be dissimilar if and only if  $A^f\bigcap B^f=\emptyset$. Otherwise, the semantic similarity between two words can be defined. For example, a trivial definition of  semantic similarity between the word $\mathfrak{A_O}$ and the word $\mathfrak{B_O}$ can be defined as: $\frac{1}{2}(\frac{|A^f\bigcap B^f|}{|A^f\bigcup B^f|}+\frac{|A_O\bigcap B_O|}{|A_O\bigcup B_O|})$.

\section{Communication}

 \par According the above analysis, self-consistent is very important for category representation. However, even when $\mathfrak{A}^{\alpha}$ is self-consistent, $\mathfrak{A}^{\alpha}$ may be something wrong. For example, \cite{Putnam1963Brains} has stated that madmen sometimes have consistent delusional systems. Why? As the final goal of a category is to help communication, different personal conceptual systems concerned on the category $\mathfrak{A}$ maybe have different representations, which results in misunderstanding, even contradiction. %unknown misunderstanding or inconsistency . %The reason is very simple, human being has a common conceptual system denoted by $L^C=\{\mathfrak{A}^C\}$.  If $\mathfrak{A}^{\alpha}$ is not consistent with $\mathfrak{A}^C$, it is still wrong.

For a communication between two men $\alpha$ and $\beta$ concerned on the category $\mathfrak{A}$,  two simple cases about the category $\mathfrak{A}$ are discussed as follows.

One case is that only one man knows the category $\mathfrak{A}$. Without loss of generality, it is assumed that the man $\alpha$ knows $\mathfrak{A}$, i.e. he knows $\mathfrak{A}^{\alpha}$ and the man $\beta$ has no idea of $\mathfrak{A}$, then the man $\beta$ must learn the knowledge about the category $\mathfrak{A}$ from $\mathfrak{A}^{\alpha}$.

In order to simplify learning, it is supposed that the $\mathfrak{A}^{\beta}$ and $\mathfrak{A}^{\alpha}$ have the same outer name  and all the  relevant conceptual systems are plain. Under such assumptions,  the man  $\beta$  needs to learn how to represent $\mathfrak{A}$ in $L^{\beta}=\{\mathfrak{A}^{\beta}\}$. Hence,  the above question can be described as follows: Let the input representation be $(O_A^{\alpha},I_A^{\alpha},\underline{A}^{\alpha},Sim_A^{\alpha})$ and the output representation be $(O_A^{\beta},I_A^{\beta},\underline{A}^{\beta},Sim_A^{\beta})$ with respect to a learning algorithm, if a subset of $(O_A^{\alpha},I_A^{\alpha})$ or $O_A^{\alpha}$ is known,  try to output $(O_A^{\beta},I_A^{\beta},\underline{A}^{\beta},Sim_A^{\beta})$, which is a standard categorization problem and has been well studied in \cite{JianYu2014Categorization},\cite{JianYu2015Categorization}.%.  $\mathfrak{A}^{\alpha}\in L^{\alpha}$outer representation of $\mathfrak{A}^{\alpha}$ is known and

The other case is that both sides in communication know the category $\mathfrak{A}$. Assume that the representation of the category $\mathfrak{A}$ for the man $\alpha$ is $(\mathfrak{A_O^{\alpha}},O_A^{\alpha},I_A^{\alpha},\mathfrak{A_I^{\alpha}},\underline{A}^{\alpha},Sim_A^{\alpha})$ and the representation of the category $\mathfrak{A}$ for the man $\beta$ is $(\mathfrak{A_O^{\beta}},O_A^{\beta},I_A^{\beta},\mathfrak{A_I^{\beta}},\underline{A}^{\beta},Sim_A^{\beta})$.   In the following, we will discuss different cases under the above assumptions.

%\subsection{General cases}

If $\mathfrak{A_O^{\alpha}}$=$\mathfrak{A_O^{\beta}}$ and $A_O^{\alpha}=A_O^{\beta}$ and $A_I^{\alpha}=A_I^{\beta}$ and $\mathfrak{A_I^{\alpha}}$=$\mathfrak{A_I}^{\beta}$ and $\underline{A}^{\alpha}=\underline{A}^{\beta}$, then it is a totally perfect communication between $\alpha$ and $\beta$ concerned on the category $\mathfrak{A}$.
 If $\mathfrak{A}^{\alpha}$ and $\mathfrak{A}^{\beta}$ are self-consistent,  it is easy to prove that the number of constraints for totally perfect communication can be greatly reduced.  Therefore, it takes for granted that categories in personal conceptual systems are self-consistent in communication.   Unfortunately, the above self-consistent conditions are  neither necessary nor sufficient for a totally perfect communication. %However, when comparing two conceptual systems, self-consistent is a natural requirement as the inner representation can not be invisible.

If $A_O^{\alpha}=A_O^{\beta}$ and $A_I^{\alpha}=A_I^{\beta}$ and $\mathfrak{A_I^{\alpha}}$=$\mathfrak{A_I^{\beta}}$ and $\underline{A}^{\alpha}=\underline{A}^{\beta}$ but $\mathfrak{A_O^{\alpha}}\neq \mathfrak{A_O^{\beta}}$, then such a category at least has two outer names when $\alpha$ and $\beta$ make no mistake.  If $\alpha$ or $\beta$ know that $\mathfrak{A_O^{\alpha}}$ and $\mathfrak{A_O^{\beta}}$ are two outer names of the category $\mathfrak{A}$, then $\alpha$ and $\beta$ still can understand each other about the category $\mathfrak{A}$, otherwise, $\alpha$ and $\beta$ can not understand each other about the category $\mathfrak{A}$.

If $\mathfrak{A_O^{\alpha}}$=$\mathfrak{A_O^{\beta}}$ and $A_O^{\alpha}=A_O^{\beta}$ and $A_I^{\alpha}=A_I^{\beta}$ and $\mathfrak{A_I^{\alpha}}$=$\mathfrak{A_I^{\beta}}$, then it is a perfect communication between $\alpha$ and $\beta$ concerned on the category $\mathfrak{A}$.

If $\mathfrak{A_O^{\alpha}}$=$\mathfrak{A_O^{\beta}}$ and $A_O^{\alpha}=A_O^{\beta}$,  then it is a semi perfect communication between the man $\alpha$ and the man $\beta$ concerned on the category $\mathfrak{A}$. Compared with perfect communication, semi perfect communication can be easily judged.  In daily life,  there are few perfect communications but more semi perfect communications. Obviously, a semi perfect communication on the category $\mathfrak{A}$ can lead to misunderstanding in some cases. If $\mathfrak{A}^{\alpha}$ and $\mathfrak{A}^{\beta}$ are self-consistent, then a semi perfect communication becomes a  perfect communication.

In practical communication, if $\overrightarrow{o_A^{\alpha}}=\overrightarrow{o_A^{\beta}}$, then it is a proper communication between the man $\alpha$ and the man $\beta$ concerned on the category $\mathfrak{A}$ and the object $o$.  In everyday functioning, a proper communication between the man $\alpha$ and the man $\beta$ concerned on the category $\mathfrak{A}$ and the object $o$ usually makes the man $\alpha$ and the man $\beta$ feel mutual understanding at least with respect to the category $\mathfrak{A}$ and the object $o$.

Sometimes, only $\mathfrak{A_O^{\alpha}}$=$\mathfrak{A_O^{\beta}}$ or $\overrightarrow{o_A^{\alpha}}=\overrightarrow{o_A^{\beta}}$ is true in communications.  In this case, misunderstanding often occurs if one side thinks communication is right.  For instance, when $\mathfrak{A_O^{\alpha}}$=$\mathfrak{A_O^{\beta}}$, $\alpha$ and $\beta$ will think they can understand each other although it is not true in many times.  For a conceptual system, a category may have several outer names, several categories may share one outer name, which brings more challenges into communications.% Hence, it causes many famous jokes.

Furthermore,when both sides know the category $\mathfrak{A}$ but only partial information is known about $\mathfrak{A}^{\alpha}$ or $\mathfrak{A}^{\beta}$, communication can still go on between the man $\alpha$ and the man $\beta$. In this case,  totally perfect communication is supposed to be true in order to make
 mutual understanding possible.  For example, $\mathfrak{A_O^{\alpha}}$ or $\mathfrak{A_O^{\beta}}$ is illustrated, communications are still available  by assuming that $\mathfrak{A^{\alpha}}=\mathfrak{A^{\beta}}$. When an object $o$  is referred to, communications still carry on by finding the category corresponding to the object $o$  by assuming that $o_A^{\alpha}=o_A^{\beta}$. Certainly, such the above cases can bring more misunderstandings and more challenges in communication as you may not see (or say) what I see (or say)\citep{Petrie1976Do}.

Usually, perfect communication is supposed to be hold in daily life in order to simplify mutual understanding. However, communication is usually not perfect and misunderstanding can not be guaranteed to be avoided,  which has resulted in so many errors, miracles, jokes, tragedies, comedies, dramas, quarrels, peace, wars and so on.

In general, if $L^{\alpha}=L^{\beta}$, then man $\alpha$ and man $\beta$ can perfectly understand each other.  If $L^{\alpha}\wedge L^{\beta}\neq \emptyset$, then man $\alpha$ and man $\beta$ are considered to have common words, otherwise, they have no common word. Usually, the larger $\frac{|L^{\alpha}\wedge L^{\beta}|}{|L^{\alpha} \vee L^{\beta}|}$, the easier man $\alpha$ and man $\beta$ communicate.

For human beings, education can make the outer representation of a category to be as same as possible for all people.   If the conceptual representation of a category is defined by a proposition, education can make the inner representation of a category to be the same for all people. Frankly speaking,  eduction can make men's conceptual systems share common words as many as possible.  When men share many common words, it can greatly reduce the dialogue cost, which is the deep reason why man prefer to using propositions to represent the conceptual representation of a category. Throughout evolution, any culture has formed enough common words so that a man can share his feeling, thought, observation, instruction, plan and imagination by common words in his community, such common words can naturally form a language.  A language reflects the common words among its users.
  %When considering this point,

\section{Words, Sentences and Knowledge}

As \cite{wittgenstein1953philosophical} has stated,  ``the meaning of a word is its use in language". Transparently, any word is used in sentences. Usually, man uses not words (phrases or symbols) but sentences to communicate with each other. Sentences describe man's feeling, thought, observation, instruction, plan and imagination through the relation between words and objects in the universe. What's relation between words and sentences?

When studying the relation between words and sentences, a category represents sentence pattern, word represents the sentence pattern name that a word is used in the corresponding sentence, an object in the discussed domain can be supposed to be a sentence. Based on the above assumption, the proposed category representation  $(\mathfrak{A_O},O_A,I_A,\mathfrak{A_I},\underline{A},Sim_A)$ can be redefined as follows:   $\forall o, o_A$ can be represented by a sentence (sometimes, an object $o$ is even a sentence. But an object $o$ is not guaranteed to be a sentence),  $\mathfrak{A_O}$ is an abstract name of a sentence pattern $\mathfrak{A}$. Consequently, $(\mathfrak{A_O},O_A,I_A,\mathfrak{A_I},\underline{A},Sim_A)$ can represent any sentence pattern.  In other words, a sentence is an concrete object representation of some sentence pattern. A sentence pattern is named by a word (phrase or symbol). %As for $\underline{A}$, it depends on how to define the corresponding sentence pattern.

In theory, when self-consistent holds for $\mathfrak{A}$, it is very important to express $\underline{A}$ in an explicit way. According to the above analysis,  $o_A$ is an instantiations of $\underline{A}$ and $\underline{A}$ is the conceptualization of $O_A$.  Hence, $\underline{A}$  can be considered as an operator as follows:
\par $\underline{A}$: $o \mapsto o_A$ such that $\underline{A}(o)=o_A$. By this way, $\underline{A}$ can be called category feature mapping.

Usually, $\underline{A}$ can be  explicitly defined.   For example,  assuming that $\underline{A}$ is defined by a predicate $P_A()$, then $o_A=P_A(o)$ as $P_A(o)$ is a statement. Under such an assumption, it is easy to know that $I_A(o_A)=Sim_A(o_A,\underline{A})\in \{0,1\}$. Therefore, the proposed category representation bridges the gap between words and propositions (statements).  In a broad sense, the proposed category representation establishes the relation between sentences and words.

It is well known that knowledge can be expressed by sentences in natural language.  Considered the relation between words and sentences, we can say that knowledge are words.  When you know all words clearly, you know all knowledge as far as man can reach.  Here, a word $\mathfrak{A_O}$ refers to $(\mathfrak{A_O},O_A,I_A,\mathfrak{A_I},\underline{A},Sim_A)$, which belongs to common words in a culture independent of any individual. By the above analysis, more words means more knowledge. A new word means some new knowledge.  Eduction helps words propagate. Man continuously creates new words to make life more advanced.  To our surprise, Stefan George  stated that ``where word breaks, nothing may be'', which also implies that no words, no knowledge.
%For man $\alpha$, a word $\mathfrak{A_O}$ refers to $(\mathfrak{A_O^{\alpha}},O_A^{\alpha},I_A^{\alpha},\mathfrak{A_I^{\alpha}},\underline{A}^{\alpha},Sim_A^{\alpha})$

%As \cite{lako1987women} has pointed out, our conceptual systems grow out of bodily experience; moreover, the core of our conceptual systems is directly grounded in perception, body movement, and experience of a physical and social character.  Therefore, the meaning of any word depends on a man's physical and social environments. Hence, for a man $\alpha$ and a category $\mathfrak{A}$, the corresponding category feature mapping should be written by $\underline{A}^{\alpha}$. Transparently, different men may have different category representations with respect to the same category $\mathfrak{A}$.

\section{Situations}

In the above analysis, a conceptual system is supposed to be independent of situations. However, a conceptual system may contain too many words and the corresponding instantiations (sentences).  Furthermore,  a man may have several conceptual systems. For example, a man may master several languages such as English, Chinese, Spanish and so on.  When will he speak English, Chinese and Spanish? If he is in Australia, he had better speak English. When he met a native farmer in Australia, he had not better talk about Beijing opera or Beijing roast duck.  Therefore, specific situation determines the optimal conceptual system and words.  Of course, a key problem for a man is to determine which conceptual system should be used at his current situation. If situations are considered as objects and conceptual systems are considered as categories, the above problem is a standard categorization question.  In the following, mathematical language will be used to describe relation between conceptual systems and situations.

In mathematics, suppose that $S=\{s(t)| t\in R\}$ is the set of situations, where $s(t)$ is the situation at time $t$. $s(t)$ consists of all objects in the universe at time $t$. In this paper, objects can be referred to any thing, including locations, individuals, things, events,properties, relations and so on. %Usually, $s(t)$

For a man $\alpha$, he only can experience his personal situations. Man $\alpha$'s situations can be represented by a set $S^{\alpha}=\{s^{\alpha}(t)| t\in R\}$, where $s^{\alpha}(t)$ is the situation at time $t$ that $\alpha$ is actually in.  In daily life, $s^{\alpha}(t)$ is very different from $s(t)$.  In general, $s^{\alpha}(t)\subset s(t)$. Moreover, no matter what conceptual system a man will choose, he actually perceives the same situation at a fixed time $t$.  Hence, it can be assumed that $s_a^{\alpha}(t)$ is the abstract situation with respect to situation $s^{\alpha}(t)$, which represents object representation that $\alpha$ actually perceives at the situation $s^{\alpha}(t)$. As \cite{Barwiseperry1999} pointed out, all the objects in $s^{\alpha}(t)$ are represented in  $s_a^{\alpha}(t)$, but some objects represented in $s_a^{\alpha}(t)$  do not belong to $s^{\alpha}(t)$. In theory, objects represented in $s_a^{\alpha}(t)$ includes objects that $\alpha$ has remembered, is experiencing. Sometimes, objects represented in $s_a^{\alpha}(t)$ are imagined or remembered by the man $\alpha$, thus they may not be objects in $s^{\alpha}(t)$.  Usually, $s^{\alpha}(t)\subseteq s_a^{\alpha}(t)$. %In some case, $s^{\alpha}(t)\subset s_a^{\alpha}(t)$

In practice, a man $\alpha$ usually has a finite conceptual systems. Assume that $\mathfrak{L}^{\alpha}=\{\mathfrak{L}^{\alpha}(t)|t\in R\}$ and $\forall t, \mathfrak{L}^{\alpha}(t)\in \{L_1^{\alpha}(t),L_2^{\alpha}(t),\cdots, L_c^{\alpha}(t)\} $, where $L^{\alpha}(t)$ is the conceptual system that $\alpha$ adopts at time $t$.   If a man adopts a unsuitable conceptual system for his current situation,  embarrassments (sometimes, dangers) may occur. Therefore, when a man $\alpha$ is in situation $s^{\alpha}(t)$, he needs to find the optimal conceptual system $\mathfrak{L}^{\alpha}(t)$.  In order to judge which conceptual system should be used, it needs to set a situation feature mapping $s_L: s_a^{\alpha}(t)\rightarrow s_L^{\alpha}(t)$, where $s_L^{\alpha}(t)$ is the situation representation corresponding to the situation $s(t)$ and the man $\alpha$.

Clearly, a conceptual system corresponds to a category. Assuming that $\underline{L}_i^{\alpha}(t)$ is the concept corresponding to the conceptual system $L_i^{\alpha}(t)$, $Sim_L^{\alpha}$  is the category similarity mapping between situations and conceptual systems. According to the study in \citep{JianYu2015Categorization}, the optimal conceptual system should be defined as $\mathfrak{L}^{\alpha}(t)$=$\widetilde{s_L^{\alpha}(t)}=\arg\max_{L_i^{\alpha}(t)}Sim_{L}^{\alpha}(s_L^{\alpha}(t),\underline{L}_i^{\alpha}(t))$.

When the optimal conceptual system $\mathfrak{L}^{\alpha}(t)$ is selected, man $\alpha$ needs to select the optimal word for the objects in the current situation $s^{\alpha}(t)$, such issues have been well studied in Section 3.  Here, the discussed domain is $O\bigcap s_a^{\alpha}(t)$ when selecting the optimal word for the current situation $s^{\alpha}(t)$.

If a man has a unique conceptual system,  he can skip the process of selecting conceptual systems when communicating with others. What a pity, a man usually has more than one conceptual system except in childhood, even though he only speaks one language.  \cite{Freud1923} had discovered that there are id, ego and superego in a personal mental life.  Such three parts generally have different (sometimes, contradictory) conceptual systems. Thus, it is doomed for man to select the optimal conceptual system to fit the specific situation.

Moreover, a situation often consists of  several sub situations and a conceptual system also has several sub conceptual systems. With respect to a special sub situation, an optimal sub conceptual system can be selected according to the same principle.  The hierarchical situation structure corresponds to similar hierarchical conceptual system structure.

More complex, $\widetilde{s_L^{\alpha}(t)}$ may not be one element, which shows that two and more conceptual systems are candidate. In this case, a candidate conceptual system usually is randomly selected.  The similar thing can happen when selecting the optimal word by inner referring operator.  Such cases inevitably brings mistakes or errors in daily life.  In theory, avoiding such cases need more refined words, conceptual systems and situation representation.

For the same situation $s(t)$, the man $\alpha$ and the man $\beta$ may have experience different situations $s^{\alpha}(t)$ and $s^{\beta}(t)$, they also uses different conceptual systems $\mathfrak{L}^{\alpha}(t)$ and $\mathfrak{L}^{\beta}(t)$. For example, the blind  and the deaf can not have the same situation experience even though they are in the same situation. However, situation's constraints on conceptual systems are the same for all men.

Situations make it is more complex  for man to understand each other. However,  situations are powerful for word sense disambiguation.  In daily life, the meaning of any pronoun is determined by situations. Without situations, it is impossible to understand what a pronoun is referred to.  Clearly, so called context is often a part of a special situation.  In some sense, man always hope to find the truth independent of situations in order to reducing the computational complexity.  Certainly, it is an impossible mission for man.  Usually, situation limitation can be weaken in some sense, but can never be totally ignored.

%When the situation is specified, object representation can be simplified by ignoring $s_L^{\alpha}(t)\bigwedge s_L^{\beta}(t)$.

\section{Denotation, Connotation and Annotation}

%When a man $\alpha$ uses a word  $\mathfrak{A_O^{\alpha}}$,  it does not always refer to an object in the physical world, sometimes it refers to one of his mental states $\underline{A}^{\alpha}$.  When referring to emotional states, Most words in any natural language have its emotional meaning.
%Any communication specifies speakers and situations.  For a given situation $s(t)$, man $\alpha$ and man $\beta$ have the same context $s^{\alpha}(t)\bigcap s^{\beta}(t)$.  According to Grice's cooperative principle, conversation should follow the maxim of quantity, which requires that 1) Make your contribution as informative as is required (for the current purposes of the exchange); 2) Do not make your contribution more informative than is required.

 For communication, if the speaker is the man $\alpha$ and the listener is the man $\beta$,  the most important thing for the listener $\beta$ is to know that ($A_I^{\alpha},\underline{A}^{\alpha}$) when man $\alpha$ said the outer name $\mathfrak{A_O^{\alpha}}$.  However, ($A_I^{\alpha},\underline{A}^{\alpha}$) cannot be observed explicitly no matter whether self-consistent is  true or not.  Therefore, sometimes the speaker and the listener can play a game together to fool others.

 In theory, it is very important to know what an element in $\mathfrak{A_O^{\alpha}}$ is referred to. Based on the study of \cite{Popper1972} and \cite{Cassirer1944}, an element in $\mathfrak{A_O^{\alpha}}$ can belong to the physical world, the mental world, and the symbolic world.

 When the elements in $A_I$ are objects in the physical world, $\mathfrak{A_I}$ is a denotation.  When the elements in $A_I$ are psychological states in the mental world, $\mathfrak{A_I}$ is a connotation.  Therefore, man can use words to refer to not only objects in the real world but also  mental states such as pains, loves, dislikes, fears, hopes, beliefs and intentions.  Generally speaking, denotation needs to rebuild its inner set in the corresponding conceptual system as its referred object is explicit, connotation must form its outer set in the corresponding conceptual system as the mental state is transparent for its owner but implicit for other men. %In this paper, mental states are called attitudes.

 When the denotations are investigated, it may be a little easy to make an agreement by referring to the objects in the physical world. When the connotations are discussed, it may be very challenge to reach a conclusion accepted by two sides as the inner referring cannot be objectively observed by two sides.   In natural language, many words (phrases or symbols) can not be absolutely denotation or connotation. For example, the word ``pig"  can make people different feelings besides its denotation. Clearly speaking, many words have some affective character.  %In other words, the discussed domain includes the physical world and the mental world.

 What's more,  words (phrases or symbols) may be neither denotation nor connotation.  In daily life, men can create new words (phrases or symbols) to express their plans, imaginations and dreams.  Before plans, imagination and dreams come true, they are only words (phrases or symbols).  In reality, most plans, imaginations and dreams have no chance of being fulfilled, hence they are not real things in the world independent of the corresponding  words (phrases or symbols).  According to \cite{Popper1972} and \cite{Cassirer1944}, words (phrases or symbols) belong to  the symbolic world. As  \cite{Cassirer1944} has pointed out,  language, myth, art, and religion are parts of the symbolic world.  In modern times, television, film, advertisement, world wide web and media are also parts of the symbolic world.

 For a self-consistent category $\mathfrak{A}$, if elements in $A_I$ do not correspond to something real in the physical world or the mental world except for the corresponding word (phrase or symbol) in the symbolic world, such a category is called an annotation.  Hence, an annotation belongs to the symbolic world.  When the discussed domain is constrained in the physical world or the mental world except for the symbolic world,  $A_I=\emptyset$ means that the corresponding category $\mathfrak{A}$ may be an literary invention but $\mathfrak{A_I}$ is an actual word  (phrase or symbol) in the symbolic world, for example, unicorn, Escher's pictures. If the discussed domain is in the symbolic world, especially in languages, $A_I\neq \emptyset$ for any category $\mathfrak{A}$ as $\mathfrak{A_O}$ can be self-referring. Generally speaking, most objects are literary invention in a fiction but indeed reflect the writer's idea, feeling and view. In theory, the symbolic world not only reflects the physical world or the mental world, but also creates something not existed in the physical world or the mental world. For example, \cite{Chomsky1956} composed his famous sentence ``colorless green ideas sleep furiously''. Here, ``colorless green ideas'' refers to nothing in the physical world, but can refer to ``colorless green ideas''  itself in language. In fact, ``colorless green ideas'' has its specific sound or its handwritten word image, ``colorless green ideas'' can refer to the corresponding sound or the corresponding handwritten word image. Any word can be used without referring a real object in the physical world or a real psychological state in the mental world. Frankly speaking, any language can be used in the symbolic world by itself, without directly referring to any thing in the physical world or the mental world.  Furthermore, many languages have been used in the world, and one language can be represented by the other language, including itself.

 If the conceptual system $L^{\alpha}$ is self-consistent,  elements in $A_I^{\alpha}$ are not objects in the physical world or the mental world, the category $\mathfrak{A^{\alpha}}$  belongs to the symbolic knowledge for the man $\alpha$. In fact, some people can use language loquaciously without referring to a real event in the physical world or the mental world such as Denyse\citep{Pinker1995}.   When elements in $A_I^{\alpha}$ are objects in the physical world or the mental world, the man really knows the category $\mathfrak{A^{\alpha}}$, $\mathfrak{A^{\alpha}}$ becomes the real knowledge for the man $\alpha$. Transparently, that $A_I^{\alpha}=\emptyset$ holds in the physical world does not mean that $A_I^{\beta}=\emptyset$ holds in the physical world. The same word may not have the same meaning for the different men.

\section{Truth, Falsity and Uncertainty}
In any conceptual system, it is very important to define what truth or falsity is.  According to \cite{LakofeJohnson1980}, truth is based on understanding. Understanding includes ``knowing yourself" and ``knowing others".  ``Knowing yourself" occurs in the same conceptual system, and ``knowing others" in two conceptual systems. In the following, we will discuss them respectively.
%Based on Section 5, 6 and 7,  understanding is based on communication between two conceptual systems.

\subsection{Knowing Yourself and Inner Truth}

Knowing yourself means that a man know his conceptual system well. In other words,  the man $\alpha$ know which category he takes to be true in his conceptual system $L^{\alpha}$.

Unfortunately, the category in the conceptual system $L^{\alpha}$ has more than the states of truth and falsity. More specifically,
for a category $\mathfrak{A^{\alpha}}$  in the conceptual system $L^{\alpha}$,  if $\exists o$ such that $\overrightarrow{o_A^{\alpha}} $ or $\widetilde{o_A^{\alpha}}$ is multiple value, then $\mathfrak{A^{\alpha}}$ is said to be uncertain, otherwise, $\mathfrak{A^{\alpha}}$ is said to be certain.  When $\mathfrak{A^{\alpha}}$ is uncertain, it is impossible to decide whether or not $\mathfrak{A^{\alpha}}$ is absolutely true or false.
If $\mathfrak{A^{\alpha}}$ can be said to be absolutely true or false, then $\forall o\in O$ ($\overrightarrow{o_A^{\alpha}} $ and $\widetilde{o_A^{\alpha}}$ are single value).

 If  $\forall o\in A_O\vee A_I (\overrightarrow{o_A^{\alpha}} $ and $\widetilde{o_A^{\alpha}}$ are single value),  $\mathfrak{A^{\alpha}}$ is said to be inner true if it is self-consistent, $\mathfrak{A^{\alpha}}$ is said to be inner false if  $\mathfrak{A_I^{\alpha}}\neq \mathfrak{A_O^{\alpha}}$ or $A_I^{\alpha}\neq A_O^{\alpha}$.  Clearly, inner true is equivalent to self-consistent.

 If $\mathfrak{A^{\alpha}}$ is uncertain, it is possible to compute whether or not $\mathfrak{A^{\alpha}}$ is  true or false in probability.  However, it still can be judged  if relation between a specific object $o$ and the category $\mathfrak{A}$ is true or false.  If $\overrightarrow{o_A^{\alpha}}=\widetilde{o_A^{\alpha}}$, then the category $\mathfrak{A^{\alpha}}$ is locally self-consistent with respect to the object $o$.  In particular, if $\overrightarrow{o_A^{\alpha}}=\widetilde{o_A^{\alpha}}$ and $\overrightarrow{o_A^{\alpha}}=\mathfrak{A_O^{\alpha}}$, then it is inner true  that the object $o$ belongs to the category  $\mathfrak{A}^{\alpha}$ .%If $\overrightarrow{o_A^{\alpha}}=\widetilde{o_A^{\alpha}}$ and $\overrightarrow{o_A^{\alpha}}\neq \mathfrak{A_O}$ and $|\overrightarrow{o_A^{\alpha}}|=1$, then it is inner false that the object $o$ belongs to the category  $\mathfrak{A}$.% Furthermore, if $|\overrightarrow{o_A^{\alpha}}|>1$,

\subsection{Knowing Others and Outer Truth}

In daily life, a conscious man $\alpha$ knows that his judgement about $\mathfrak{A}$ may not be objectively true even if $\mathfrak{A^{\alpha}}$ is inner true.  Usually, a conscious man needs compare his judgement with others.  Let us consider the simplest case.  Man $\alpha$ only needs to consider the opinion of man $\beta$ with respect to $\mathfrak{A}$. If $\mathfrak{A_O^{\alpha}}$=$\mathfrak{A_O^{\beta}}$ and $A_O^{\alpha}=A_O^{\beta}$, then man $\alpha$ thinks that $\mathfrak{A^{\alpha}}$ is outer true with respect to man $\beta$. If $\mathfrak{A_O^{\alpha}}\neq\mathfrak{A_O^{\beta}}$ or $A_O^{\alpha}\neq A_O^{\beta}$, then man $\alpha$ thinks that $\mathfrak{A^{\alpha}}$ is outer false with respect to man $\beta$. In practice, the above condition is still very demanding for man $\alpha$ to judge whether or not $\mathfrak{A^{\alpha}}$ is outer true with respect to man $\beta$. Therefore, the condition of the outer truth needs further simplified with respect to a category.

It is very difficult to get $A_O^{\alpha}$ and $A_O^{\beta}$ in general cases, but it is easier to consider  an object in $A_O^{\alpha}$ or $A_I^{\alpha}$. When outer referring operator is supposed to be single value, man $\alpha$ thinks that it is outer true with respect to man $\beta$ that the object $o$ belongs to $\mathfrak{A^{\alpha}}$ if $\overrightarrow{o_A^{\alpha}}=\overrightarrow{o_A^{\beta}}=\mathfrak{A_O^{\alpha}}$
, and  $\alpha$ thinks that it is outer false with respect to man $\beta$ that the object $o$ belongs to $\mathfrak{A^{\alpha}}$ if $\overrightarrow{o_A^{\alpha}}=\mathfrak{A_O^{\alpha}}\neq \overrightarrow{o_A^{\beta}}$.  Certainly, it is easier to judge whether or not $\overrightarrow{o_A^{\alpha}}=\overrightarrow{o_A^{\beta}}$ holds than  $\mathfrak{A_O^{\alpha}}$=$\mathfrak{A_O^{\beta}}$ and $A_O^{\alpha}=A_O^{\beta}$.

In particular, if man $\beta$ is supposed to be an oracle, man $\alpha$ thinks that $\mathfrak{A^{\alpha}}$ is outer true if $\mathfrak{A_O^{\alpha}}$=$\mathfrak{A_O^{\beta}}$ and $A_O^{\alpha}=A_O^{\beta}$. Man $\alpha$ thinks that it is outer true that the object $o$ belongs to $\mathfrak{A^{\alpha}}$ if $\overrightarrow{o_A^{\alpha}}=\overrightarrow{o_A^{\beta}}=\mathfrak{A_O^{\alpha}}$ and man $\alpha$ thinks that it is outer false  that the object $o$ belongs to $\mathfrak{A^{\alpha}}$ if $\overrightarrow{o_A^{\alpha}}=\mathfrak{A_O^{\alpha}}\neq \overrightarrow{o_A^{\beta}}$.  Transparently, the inner truth is not equivalent to the outer truth. Moreover, the inner truth is independent of the outer truth.

\subsection{Truth, Falsity, Uncertainty}

Transparently, inner truth and outer truth are not guaranteed to be true. In logic, it is always assumed that absolute truth is independent of different conceptual systems.
In order to achieve this aim, a natural assumption is that self-consistent is self evident and man $\beta$ must be an oracle. Even when $\mathfrak{A^{\alpha}}$ is self-consistent, $\mathfrak{A^{\alpha}}$ still depends on man $\alpha$. In order to make category representation impersonal, a simple assumption is that $\alpha=\beta$.
 %If $\mathfrak{A^{\alpha}}$ belongs to common words of a culture, then $\mathfrak{A^{\alpha}}$ will not depend on man $\alpha$. In the following, a category $\mathfrak{A}$ is independent of different conceptual systems.

Under the above assumption, it is easy to judge whether or not $\mathfrak{A}$ is true.  As a matter of fact, if $\mathfrak{A}$ is inner true, then $\mathfrak{A}$ is true.  If $\mathfrak{A}$ is inner false, then $\mathfrak{A}$ is false. In common sense, if $\overrightarrow{o_A}=\widetilde{o_A}=\mathfrak{A_O}$, it is said to be true that the object $o$ belongs to $\mathfrak{A}$; if $\overrightarrow{o_A}\neq\widetilde{o_A}$ or $\overrightarrow{o_A} \neq\mathfrak{A_O}$, it is said to be false that the object $o$ belongs to $\mathfrak{A}$.    Certainly, when making the above judgments,  the presupposition is made as follows:  $\forall o\in O$ (its object representation $o_A$ is independent of any observer) and $\forall o\in (A_O\vee A_I)$($\overrightarrow{o_A}$ and $\widetilde{o_A}$ are single value).

Under the above assumptions and presuppositions, if  $\mathfrak{A}$ is true, then it is  not only inner true but also outer true for anyone. If  $\exists o\in O$(($\overrightarrow{o_A}$ or $\widetilde{o_A}$ is multiple value) and ($\mathfrak{A_O}\in \overrightarrow{o_A}$ or $\mathfrak{A_I}\in \widetilde{o_A}$)), $\mathfrak{A}$ is uncertain. In particular, if the object $o\in O$(($\overrightarrow{o_A}$ or $\widetilde{o_A}$ is multiple value)and ($\mathfrak{A_O}\in \overrightarrow{o_A}$ or $\mathfrak{A_I}\in \widetilde{o_A}$)), then it is uncertain that the object $o$ belongs to the category $\mathfrak{A}$ without specification.

%According to analysis in section 6, suppose that $\underline{A}=P_A()$, $o_A=P_A(o)$, $I_A(o_A)=Sim_A(o_A,\underline{A})\in \{0,1\}$ where $P_A()$  is a predicate.  As $\overrightarrow{o_A}=\mathfrak{A_O}$ implies $I_A(o_A)=1$, and $I_A(o_A)=1$ clearly means that $P_A(o)$ is true, it can be proved that $\overrightarrow{o_A}=\mathfrak{A_O}$ means that $P_A(o)$ is true.  Set $P_A()=$ belongs to $\mathfrak{A}$, then the above definition of truth follows that if $P_A(o)$ is true, $P_A(o)$ is true. Therefore, the

Obviously, the above assumption about truth is too strong to be satisfied in daily life.  Obviously, it is little chance for a man to be an oracle. Therefore, truth in daily life can not be defined independent of different conceptual systems.   In common sense,  $\mathfrak{A^{\alpha}}$ is empirically true with respect to man $\alpha$ and man $\beta$ if and only if $\mathfrak{A^{\alpha}}$ and $\mathfrak{A^{\beta}}$ are inner true and $\mathfrak{A^{\alpha}}$ is outer true with respect to man $\beta$.  Man $\alpha$ thinks that $\mathfrak{A^{\alpha}}$ is empirically true with respect to man $\beta$ if and only if $\mathfrak{A^{\alpha}}$ is inner true and $\mathfrak{A^{\alpha}}$ is outer true with respect to man $\beta$.  More specifically, it is empirically true with respect to man $\alpha$ and man $\beta$ that the object $o$ belongs to the category $\mathfrak{A}$ if and only if  $\overrightarrow{o_A^{\alpha}}=\overrightarrow{o_A^{\beta}}=\mathfrak{A_O^{\alpha}}$ and $\overrightarrow{o_A^{\alpha}}=\widetilde{o_A^{\alpha}}$ and $\overrightarrow{o_A^{\beta}}=\widetilde{o_A^{\beta}}$. If and only if  $\overrightarrow{o_A^{\alpha}}=\overrightarrow{o_A^{\beta}}=\mathfrak{A_O^{\alpha}}$ and $\overrightarrow{o_A^{\alpha}}=\widetilde{o_A^{\alpha}}$, man $\alpha$ thinks it is empirically true with respect to man $\beta$ that the object $o$ belongs to the category $\mathfrak{A}$.

 In daily life, inner truth, outer truth and empirical truth is more useful than truth. Sometimes, outer truth, inner truth and empirical truth are confused with truth in daily life.  However, there are great differences among inner truth, outer truth, empirical truth and truth.  Sometimes, inner truth holds but outer truth does not hold. A famous example in \cite{Betrandrusell1946} is that the lunatic who believes that he is a poached egg is to be condemned solely on the ground that he is in a minority, or rather on the ground that the government does not agree with him.  Certainly, the category ``the poached egg'' is self-consistent with respect to such an lunatic, therefore, it is inner true with respect to such an lunatic, but it can not be accepted as outer truth by others that lunatic has the right knowledge about ``the poached egg''.   Therefore,  the inner truth is not guaranteed to be the outer truth. Similarly, outer truth does not guarantee inner truth.  The emperor's new clothes clearly illustrated the difference between outer truth and inner truth.  History of science has illustrated many times that empirical truth is equivalent to truth.

More specifically, when talking about connotations, inner truth is more important than outer truth; when referring to denotations, outer truth plays more important role than inner truth; when discussing annotations, outer truth have the same influence as inner truth. In everyday life, man should carefully distinguish when inner truth, outer truth, empirical truth and truth holds.

%More accurately, we should carefully distinguish inner truth, outer truth and truth in everyday life. %Generally speaking, truth seems not existed in daily life but

\section{Related Work}

In this paper, we will partially survey some papers in the literature related to communication and truth.

%\subsection{Category Theories}

%The proposed semantic set $(\mathfrak{A_O},O_A,I_A,\mathfrak{A_I},\underline{A},Sim_A)$ can lead to many concepts in the literature. In the following, we will give a brief survey on this issue.
%Set that $I_A \in \{0,1\}$, $\underline{A}$ is a predicate letter $P_A()$ and $Sim_A $ is the valuation of $P_A()$.  If $I_A=Sim_A$ and $\mathfrak{A_O}=\mathfrak{A_I}$, then it is easy to know that a classic set with name $\mathfrak{A_O}$ becomes a special case of a semantic set.

%Let $\underline{A}$ be the prototype corresponding to the category $\mathfrak{A}$. If $I_A=Sim_A$ and $\mathfrak{A_O}=\mathfrak{A_I}$, then it is easy to know that the proposed semantic set is degenerated into prototype theory \citep{rosch1978principles}.

%Let $\underline{A}$  be a set consisting of multiple exemplars.  If $I_A=Sim_A$ and $\mathfrak{A_O}=\mathfrak{A_I}$, then it is easy to know that the proposed semantic set is degenerated into exemplar theory \citep{medin1978context}.

%As for knowledge theory, a category  is part of a general knowledge about the world \citep{murphy1985role}.

%\subsection{Situation and Attitude}

\subsection{Grice's Cooperative Principle}

\par For communication, \cite{Grice1975} presented cooperative principle as follows: make your conversational contribution such as is required, at the stage at which it occurs, by the accepted purpose or direction of the talk exchange in which you are engaged.
\par \cite{Grice1975} thought that cooperative principle can yield four maxims, including quantity, quality, relation and manner as follows:\\
\textbf{Quantity: }Make your contribution as informative as is required (for the current purposes of the exchange); do not make your contribution more informative than is required. \\
\textbf{Quality:} Do not say what you believe to be false; do not say that for which you lack adequate evidence.\\
\textbf{Relation:} Be relevant.\\
\textbf{Manner:} Avoid obscurity of expression; avoid ambiguity; be brief (avoid unnecessary prolixity); be orderly.

Transparently, the listener must understand the speaker's meaning in order to follow the cooperative principle during conversation.  When the accepted purpose or direction of the talk exchange between man $\alpha$ and man $\beta$ is about the category $\mathfrak{A}$, if  man $\alpha$ says $\mathfrak{A_O^{\alpha}}$, man $\beta$ should assume that $\mathfrak{A_O^{\alpha}}$=$\mathfrak{A_O^{\beta}}$ and $A_O^{\alpha}=A_O^{\beta}$ and $A_I^{\alpha}=A_I^{\beta}$ and $\mathfrak{A_I^{\alpha}}$=$\mathfrak{A_I}^{\beta}$ and $\underline{A}^{\alpha}=\underline{A}^{\beta}$. % Furthermore, assume that situation for man $\beta$ is $s(t)^{\beta}$, then the abstract situation for man $\beta$ is $s(t)_L^{\beta}\bigwedge \mathfrak{A_O^{\beta}} $,

Under such an assumption, maxim of quantity requires that the man $\alpha$ refers to the object $o$ with respect to the category $\mathfrak{A}$, the man $\beta$ should provide the information $o_A^{\beta}-o_A^{\alpha}$.  For maxim of quality, its first condition in requires categories in personal conceptual systems are self-consistent, which means that $A_O^{\alpha}=A_I^{\alpha}$ and $\mathfrak{A_O^{\alpha}}=\mathfrak{A_I^{\alpha}}$; its second condition implies $A_O^{\alpha}\neq \emptyset$. The same requirements for the man $\beta$ is also true.  Maxim of relation and maxim of manner do something with not only relation between situations and conceptual systems but also totally perfect  communication condition, which require that people must choose the right conceptual system according to the specific situation, otherwise, maxim of relation and maxim of manner may be violated in great probability.

As pointed out in Section 5, totally perfect communication condition does not ask for self-consistent, which implies that cooperative principle is not fully consistent with maxim of quality. In daily life,  many stories demonstrate this point.  If totally perfect communication condition is satisfied, cooperative principle may be obeyed.  If totally perfect communication condition is not satisfied, cooperative principle can not be obeyed.   \cite{Pinker2007} has cited an example as follows: In an episode of \emph{Seinfeld}, George is asked by his date if he would like to come up for coffee. He declines, explaining that caffeine keeps him up at night. Later he slaps his forehead and realizes, ```Coffee' does not mean coffee! `Coffee' means sex". In this example, George and his date clearly did not satisfy the totally perfect communication condition, and George's date also violated the self-consistent condition.

In summary, Grice's cooperative principle is just a roughly description about self-consistent and  the relation between word and the corresponding category feature representation, totally perfect communication condition and the proposed relation between situation and conceptual system offer more powerful explanation for communication than Grice's cooperative principle.
%However, according to the analysis, the man $\beta$ should state the information relevant to the category $\mathfrak{A}$ maxim of quantity implies that the man $\beta$ should state the information  $\mathfrak{A^{\beta}}-(s_L^{\beta}(t)\bigwedge s_L^{\alpha}(t))$ by assuming that abstract situation for man $\alpha$ is $s_L^{\alpha}(t)$ and the abstract situation for man $\beta$ is $s_L^{\beta}(t)$.

\subsection{On Truth}

In the literature, many papers have devoted to defining truth such as correspondence theory, coherence theory, experientialist theory and so on \citep{LakofeJohnson1980}. %In this section, we only discuss the relations between the proposed result

The most famous definition of  truth is originally based on the well known words of \cite{Aristotle1908}: ``To say of what is that it is not, or what is not that it is, is false, while to say of what is that it is , or what is not that it is not, is true".  Such statement has been translated by \cite{Tarski1956} as  `` a true sentence is one which says that the state of affairs is so and so, and the state of affairs indeed is so and so".  Then it formulated the classic Tarski truth definition, which can be intuitively interpreted as  `` Snow is white" is true if and only if snow is white.

However, the well known words of \cite{Aristotle1908} about truth can be interpreted differently from classic Tarski truth definition.  In some sense, the proposed definitions of inner truth, outer truth, empirical truth and truth  also follow Aristotle truth definition.  Therefore, the study of truth in this paper is also consistent with correspondence theory.  

If considering self-consistent as self understanding, and semi perfect communication as mutual understanding, then the definition of inner truth, outer truth, empirical truth and truth are based on understanding.  So inner truth, outer truth, empirical truth and truth are consistent with \cite{LakofeJohnson1980}, which says that ``we understand a statement of being true in a given situation when our understanding of the statement fits our understanding of the situation closely enough for our purposes".  Therefore, the proposed study of truth has all the characteristics of \cite{LakofeJohnson1980}'s experientialist theory of truth, in other words,    the study of truth in this paper has some elements of a coherence theory and a pragmatic theory.

\section{Discussion and Conclusions}

When ignoring category name,  \cite{JianYu2015Categorization} has presented a category representation. However, everything has a name, as \cite{Cassirer1944} pointed out. In this paper, a new approach to representing words (phrases or symbols) is given by considering the category name. This proposed category representation solves the drawbacks of the classical category representation, it can be taken as a generalization of a classic set. %In practice, the proposed category representation can be used as pattern representation,

\cite{Betrandrusell1950} has stated that ``some words have two non-verbal uses, (a) as indicating objects (b) as expressing states of mind''.  In this paper, such observations are extended to  all categories. The proposed category representation has three parts: one is about linguistic (word, phrase or symbol), one is about action (outer set), one is about idea (inner representation). According to \cite{Popper1972} , words (phrases or symbols) belong to the third world (symbolic world), outer set belongs to the first world (physical world) , and inner representation belongs to the second world (mental world). Hence, the category name can be categorized into denotation, connotation and annotation with respect to a conceptual system. Denotation refers to real object in the physical world. Connotation refers to psychological state in the mental world.   Annotation corresponds to object in the symbolic world but without referring objects in the physical world or the mental world. Sometimes, man can turn some annotation into reality. Under such an case,  an annotation becomes a denotation or connotation.  %In the words of \cite{LakofeJohnson1980},  words belong to culture, outer set relates to the physical world, inner representation  man.

According the above analysis, the proposed category representation establishes the relations among three worlds  \cite{Popper1972}. In daily life, every part of the category representation can independently form one conceptual system, i.e. say one thing, do another, think differently from saying and doing.  For instance,  men can smartly construct temporary conceptual systems to fit the local environmental requirements, such as actors, pretenders, spies, translators, etc. Generally speaking, man's conceptual systems vary with times and environments.  Man continuously changes his own conceptual systems through learning from others or the world.

When a man has several conceptual systems, he usually selects the optimal conceptual systems and words with respect to the specific situation. Situations determine the adoption and evolution of conceptual systems. However, situations are very complex.  Usually, $s^{\alpha}(t)$ is different from $s_L^{\alpha}(t)$. And the objects in $s^{\alpha}(t)$ can be observed by other men in the time $t$,  and the objects represented in $s_L^{\alpha}(t)$ may be not observed by other men in the time $t$ but can be imagined or remembered by the man $\alpha$.

%For example, what's the key differences for name decoding mapping and name encoding mapping between pretenders and translators?

Based on the proposed category representation, self-consistent and perfect communication are defined. Naturally, self-consistent is another version of categorization equivalency axiom proposed by \cite{JianYu2015Categorization} and totally perfect communication reinterprets the uniqueness axiom of category representation.   Self-consistent greatly reduces understanding complexity and cognitive effort  because of representational simplicity. Therefore, self-consistent can be called the economy assumption of category representation.

In this paper, every man has his own conceptual system, which is consistent with the experientialist theory \citep{LakofeJohnson1980}.

When two conceptual systems communicate, self-consistent is the lowest cost requirement as the inner representation can not be observable. When $\mathfrak{A}^{\alpha}$ is self-consistent, the man $\alpha$ is said to tell his truth (be honest) with respect to the category $\mathfrak{A}$, otherwise,  the man $\alpha$ is lying with respect to the category $\mathfrak{A}$.  The above analysis offers a clear explanation that prohibitions again lying is basic moral rule is social life just because lying can greatly increase understanding complexity and social cost.  However, when $\mathfrak{A}^{\alpha}$ is self-consistent, the man $\alpha$ cannot be guaranteed to be objectively true with respect to the category $\mathfrak{A}$.

When self-consistent is false but the inner referring set equals the outer referring set, it is worth further investigation how to construct name decoding mapping and name encoding mapping in a conceptual system.  Translation is such a typical task.  However, an ordinary person  usually has no mapping between the inner name and the outer name in his conceptual system.   Obviously, when self-consistent is not true for a category, man is considered to tell a lie. For instance, \cite{Harari2011} stated that most Christians did not imitate Christ, most Buddhists failed to follow Buddha, and most Confucians would have caused Confucius a temper tantrum. Many researches have been devoted to studying lying, such as \cite{DePaulo1996,Heyman2009} and so on.

When self-consistent is supposed to be true for category representation,  relations between words are mathematically defined such as homonymy, polysemy, synonymy, hyponymy, antonymy, metaphor and metonymy, similarity.  An interesting conclusion is that  every word has more than two meanings, which latently supports that every man may have more than one conceptual systems.

In practice, different conceptual systems often have different influences, which can result in more complex practical communication \citep{YoungFitzgerald2011}. In this paper, all the analysis ignores that the influence of $L^{\alpha}$ is different from that of $L^{\beta}$, in other words, all conceptual systems are considered to have the same influence. Even under such an assumption, only two simple cases about communications are discussed in this paper. One case is about learning. The other case is about understanding  about the same category when two sides of communication both think that they know such a category. The conditions for perfect communication and relevant cases are presented.  More complex cases such as both communication sides only know something about $\mathfrak{A}$ need to further study. Theoretically,  all the above research results can be generalized into  any two agents with their own conceptual systems. Such two agents can be any two sides, maybe two men,  two robots, one robot and one man, one man and one book, and so on. In particular, if one side in communication is supposed to be absolutely true, then the other side is considered to be true or false depending on whether the proper communication holds or not. Furthermore, when one conceptual system has superpower influence,  other conceptual systems may have been assimilated continuously or suppressed totally.  In addition, only word understanding in single turn dialogue is studied in this paper, it is natural to investigate how to understand multi-turn dialogue in the future.
%For a robot, if he owns several conceptual systems, this issue may
%how to select proper conceptual systems is a challenge

The proposed semantic set also establishes the relation between sentences and words, which clearly shows that sentences are instantiations of the corresponding words and words are conceptualization of the corresponding sentences.  Therefore, a surprising conclusion can be made: words are all knowledge, i.e. there is no knowledge without words.  Here, words include spoken words, body words, musical words, etc.  When a new word is created, some new knowledge is obtained. When some words are obsolete, their corresponding knowledge is also out of date. When some words are updated, relevant knowledge is also renewed. Therefore, words are evolved with situations, knowledge are also evolved with situations. Words and knowledge also follow the evolutionary principle:  survival of the fittest.

Last but not least, the proposed semantic set also introduces new definition of truth and analyzes the difference between inner truth, outer truth, empirical truth and truth. In daily life, inner truth, outer truth, empirical truth may be called truth.  However, there are a great gap between inner truth, outer truth, empirical truth and truth. Different men may have different inner truth, outer truth, empirical truth about the same category.  Strictly speaking, inner truth, outer truth, empirical truth are constrained by specific situations, truth does not depend on specific situations in some sense.

%In summary, the proposed semantic set provides a tool to represent the computation in natural language processing.

% Acknowledgements should go at the end, before appendices and references

\subsubsection*{Acknowledgements}
This work was partially supported by the NSFC grant (61370129).%, Ph.D Programs Foundation of Ministry of Education of China (20120009110006).
%Use unnumbered third level headings for the acknowledgements title.
%All acknowledgements go at the end of the paper.

%\subsubsection*{References}
\bibliographystyle{apa}
\bibliography{bibfile}

\appendix

%\end{CJK*}
\end{document}